%% file: main.tex
\documentclass[letterpaper, 10 pt, conference]{ieeeconf}  

\IEEEoverridecommandlockouts                              
\include{package}
\include{header}

\include{notation}
\usepackage{mysymbol}
\input{macros.tex}

\newtheorem{remark}{\textbf{Remark}}

\title{\LARGE \bf
Reactive Informative Planning for Mobile Manipulation Tasks \\under Sensing and Environmental Uncertainty
}

\author{Mariliza Tzes$^{1}$, Vasileios Vasilopoulos$^{2}$, Yiannis Kantaros$^{3}$ and George J. Pappas$^{1}$
\thanks{$^{1}$GRASP Lab, University of Pennsylvania, Philadelphia, PA 19104, USA, {\tt\small \{mtzes, pappasg\}@seas.upenn.edu}.}%
\thanks{$^{2}$Computer Science and Artificial Intelligence Laboratory (CSAIL), MIT, Cambridge, MA 02139, \texttt{vvasilo@mit.edu}.}%
\thanks{$^{3}$Department of Electrical and Systems Engineering, Washington University in St. Louis, St. Louis, MO, 63112, \texttt{ioannisk@wustl.edu}.}%
\thanks{This work was supported by AFOSR grant FA9550-19-1-0265.}
}

\begin{document}

\maketitle
\thispagestyle{empty}
\pagestyle{empty}

\begin{abstract}
In this paper we address mobile manipulation planning problems in the presence of sensing and environmental uncertainty. In particular, we consider mobile sensing manipulators operating in environments with unknown geometry and uncertain movable objects, while being responsible for accomplishing tasks requiring grasping and releasing objects in a logical fashion. Existing algorithms either do not scale well or neglect sensing and/or environmental uncertainty. To face these challenges, we propose a hybrid control architecture, where a symbolic controller generates high-level manipulation commands (e.g., grasp an object) based on environmental feedback, an informative planner designs paths to actively decrease the uncertainty of objects of interest, and a continuous reactive controller tracks the sparse waypoints comprising the informative paths while avoiding a priori unknown obstacles. The overall architecture can handle environmental and sensing uncertainty online, as the robot explores its workspace. Using numerical simulations, we show that the proposed architecture can handle tasks of increased complexity while responding to unanticipated adverse configurations.
\end{abstract}


\input{1-introduction.tex}
\input{2-problem-formulation}
\input{3-ltl-planner-icra}
\input{5-informative-sampling-based-planner}
\input{6-reactive-planner}
\input{7-simulations}
\input{8-conclusion}







\bibliographystyle{IEEEtran}
\bibliography{IEEEabrv,references, YK_bib, mariliza_bib}

\end{document}

%% file: package.tex
\usepackage{enumerate}
\usepackage[cmex10]{amsmath} 
\usepackage{amssymb} 
\usepackage{amsfonts}

\usepackage{amsthm}
\usepackage{amsbsy}
\usepackage{bm}
\usepackage{mathtools}

\usepackage{graphicx}
\usepackage{graphics} 
\usepackage{epsfig} 
\usepackage{accents}
\usepackage{algorithm}
\usepackage{algpseudocode}
\usepackage{subcaption}

\usepackage[font=small,skip=1pt]{caption}
\algdef{SE}[DOWHILE]{Do}{doWhile}{\algorithmicdo}[1]{\algorithmicwhile\ #1}%

\usepackage{color}
\makeatletter
\let\NAT@parse\undefined
\makeatother

\usepackage{cite}

\usepackage{nohyperref}
\usepackage{url}
\usepackage{comment}

\usepackage[compact]{titlesec}
\titlespacing{\section}{0pt}{2ex}{1ex}
\titlespacing{\subsection}{0pt}{1ex}{0ex}
\titlespacing{\subsubsection}{0pt}{0.5ex}{0ex}

%% file: header.tex
	\newtheoremstyle{myplain}
	  {}
	  {}
	  {\itshape}
	  {}
	  {\bfseries}
	  {}
	  {5pt plus 1pt minus 1pt}
	  {}

	\newtheoremstyle{mydefinition}
	  {}
	  {}
	  {\normalfont}
	  {}
	  {\bfseries}
	  {}
	  {5pt plus 1pt minus 1pt}
	  {}
  
	\theoremstyle{myplain}
    \newtheorem{assumption}{Assumption}

	\newtheorem{theorem}{Theorem}

	\theoremstyle{mydefinition}

   \newcommand{\field}[1]{\mathbb{#1}}
   \newcommand{\reals}{\field{R}}

\algdef{SE}[DOWHILE]{Do}{doWhile}{\algorithmicdo}[1]{\algorithmicwhile\ #1}%


%% file: macros.tex
\newcommand{\robotradius}{r} 

\newcommand{\robotposition}{\mathbf{x}} 

\newcommand{\robotpositionunicycle}{\overline{\mathbf{x}}} 
\newcommand{\robotorientation}{\psi} 

\newcommand{\freespace}{\mathcal{F}} 

\newcommand{\movableobject}{\tilde{M}} 

\newcommand{\movableobjectsetdilated}{\mathcal{M}} 
\newcommand{\movableobjectcardinality}{N_M} 
\newcommand{\movableobjectradius}[1]{\rho_{#1}} 

\newcommand{\controlfullyactuated}{\mathbf{u}} 
\newcommand{\controlunicycle}{\overline{\controlfullyactuated}} 
\newcommand{\linearinput}{v} 
\newcommand{\angularinput}{\omega} 

\newcommand{\ballclosure}[2]{\overline{\mathsf{B}(#1,#2)}}

%% file: 1-introduction.tex
\section{INTRODUCTION}
\label{sec:introduction}

Task and motion planning (TAMP) has emerged as the `backbone' of robotic manipulation, widely seen in industrial and service applications. Specifically, rearrangement planning has recently received increasing attention~\cite{garrett_ijrr_2018, lee2019efficient, toussaint2015logic, 6906922, li2021reactive, adu2021probabilistic}, but limitations still exist due to the provable NP-hardness of the overall problem~\cite{wilfong1991motion}.
One of the key challenges in the rearrangement planning scenarios is the handling of uncertainties over the task domain. Consider the scenario where the \textit{mobile} robotic manipulator is assigned to grasp an object of unknown location while avoiding unexpected conditions of the environment, e.g. unfamiliar obstacles.

\begin{figure}[t]
 \captionsetup{width=\linewidth,font=footnotesize}
         \centering
         \includegraphics[width=0.3\textwidth]{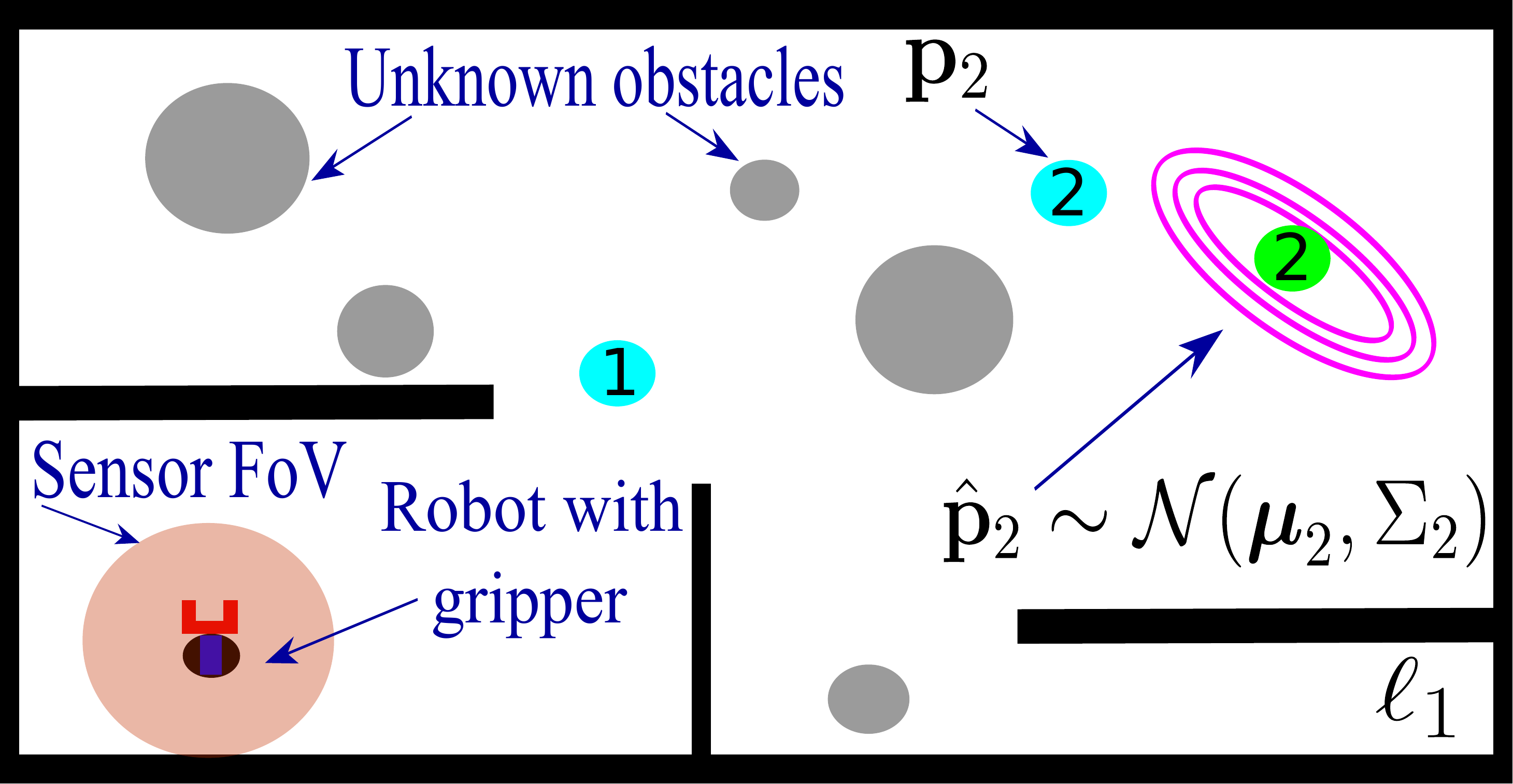}
         \setlength{\belowcaptionskip}{-22 pt}
         \caption{An illustration of the uncertainty over the task domain. The robot is equipped with a gripper and a limited range onboard sensor (orange) for localizing the uncertain movable objects (cyan) and avoid unknown, unanticipated obstacles (grey) while navigating in a partially known environment (black). }
     \label{fig:front_figure}
\end{figure}

In this work, we propose an architecture for addressing \textit{mobile} manipulation task planning problems in the presence of environmental and sensing uncertainty. In particular, we consider mobile manipulators equipped with noisy sensors being responsible for accomplishing high level manipulation tasks, captured by Linear Temporal Logic (LTL) formulas, in environments with unknown geometry and movable objects that are located at uncertain positions. In Fig. \ref{fig:front_figure} we illustrate an example of the task domain. To address such problems, we propose a novel hybrid architecture, seen in Fig. \ref{fig:system}, which can handle unanticipated conditions in the robot's workspace. Particularly, the proposed method consists of a symbolic controller generating high-level manipulation commands (e.g., grasp an object), an informative planner generating sequences of (sparse) waypoints to actively decrease the uncertainty of the objects associated with the symbolic commands, and a reactive controller to follow the informative waypoints while avoiding a priori unknown obstacles. The transition from the symbolic to the continuous reactive controller is online and can respond to unanticipated conditions such as movable objects prohibiting task accomplishment by pushing them out of the way (see \cite{vasilopoulos2021reactive}). Despite the problem complexity, our architecture is provable complete under specific conditions, and scales well with the task complexity.

\subsection{Mobile Manipulation under Uncertainty}
Most of the existing works focus on \textit{known} environments \cite{srivastava2014combined, he2015towards}, making them applicable only to very specific problem instances. Recent works on uncertain environments \cite{driess2021learning, 9036915, 8462863} propose deep learning approaches that design visually guided rearrangement planning algorithms. In \cite{9036915}, the authors introduce a visual state prediction engine, that predicts a workspace state offline. Common in these works is that the robot's camera has access to the entire workspace, decodes environmental uncertainties offline, and executes the manipulation task based on that decoded environment.

Another suggested way to approach rearrangement planning under uncertainty is to model the problem as a \textit{partially observable Markov decision process} (POMDP), where the robot holds a belief about the state of the workspace and intentionally selects
actions that reduce its uncertainty about the world \cite{garrett2020online, kaelbling2013integrated}. Garrett et. al. \cite{garrett2020online} the  An interesting approach is proposed in \cite{migimatsu2020object}, where a motion planner is defined in object-centric coordinates, enabling the derivation of a controller that can react to (small) perturbations. 

Our scheme differs from the existing literature in that it formally defines the uncertainty over the task domain and incorporates the \textit{active sensing} part, where the algorithm provides directions on to where the \textit{mobile} robotic manipulator should take measurements \textit{on-the-fly} to reduce its uncertainty, while navigating through unknown obstacles to satisfy a complex manipulation task encoded in an LTL formula.

\subsection{Reactive Temporal Logic Planning}
Examples of reactive temporal logic planning algorithms in partially unknown environments have been developed in \cite{guo2013revising,guo2015multi,maly2013iterative,lahijanian2016iterative,livingston2012backtracking,livingston2013patching,kress2009temporal,alonso2018reactive, kantaros2019optimal,kantaros2020reactive}. 
Reactive to LTL specifications planning algorithms are proposed in \cite{kress2009temporal,alonso2018reactive}, as well. Specifically, in \cite{kress2009temporal,alonso2018reactive} the robot reacts to the environment while the task specification captures this reactivity. Correctness of these algorithms is guaranteed if the robot operates in an environment that satisfies the assumptions that were explicitly modeled in the task specification. Unlike our approach, common in all these works is that they assume perfect sensors while often relying on discrete abstractions of the robot dynamics  \cite{belta2005discrete,pola2008approximately}. Additionally, the above works neglect active interaction with the environment to satisfy the logic specification. 

\subsection{Contribution} 
This paper proposes the first planning and control architecture for mobile manipulation tasks in the presence of environmental and sensing uncertainty. To the best of our knowledge, the most relevant work to the one presented here is the recent work by the authors  \cite{vasilopoulos2021reactive}, which, unlike this work, considers perfect sensors and known movable objects. The symbolic and reactive controllers maintain respective correctness and collision avoidance guarantees as in \cite{vasilopoulos2021reactive}, and in this work we present the conditions for the informative planner's correctness. Additionally, we provide a variety of simulation examples that illustrate the efficacy of the proposed algorithm for accomplishing complex manipulation tasks in unknown environments. 


%% file: 2-problem-formulation.tex
\section{PROBLEM DESCRIPTION}
\label{sec:problem_formulation}



We consider a first-order, nonholonomically-constrained, disk-shaped robot of radius $r \in \reals_{>0}$ that resides in a closed, compact, polygonal and typically non-convex workspace $\Omega \subset \reals^2$.
The robot's rigid placement is denoted by $\robotpositionunicycle(t) := (\robotposition(t),\robotorientation(t)) \in \mathbb{R}^2 \times S^1$ where $\robotposition(t) \in \reals^2$ and $\robotorientation(t) \in S^1$ are the robot's position and orientation respectively at time $t$ and its input vector $\controlunicycle(t):=(\linearinput(t),\angularinput(t))$ consists of a fore-aft and an angular velocity command. The robot's rigid placement $\bar{\mathbf{x}}(t)$ is assumed to be perfectly known at each time $t$.

The workspace $\Omega$ contains a finite collection of (i) disk-shaped \textit{movable objects} denoted by $\tilde{\mathcal{M}} \coloneqq \{ \tilde{\mathcal{M}}_i \}_{i \in \{1,\dots, N_M\}}$ with a vector of radii $(\movableobjectradius{1}, \ldots, \movableobjectradius{\movableobjectcardinality}) \in \reals^{\movableobjectcardinality}$ and (ii) \textit{disjoint obstacles} of unknown number and placement, denoted by $\tilde{\mathcal{O}}$. A subset $\tilde{\mathcal{P}} \subseteq \tilde{\mathcal{O}}$ of these obstacles are assumed to have a recognizable polygonal geometry, that the robot can instantly identify and localize as in \cite{vasilopoulos_pavlakos_bowman_caporale_daniilidis_pappas_koditschek_2020}, but a completely random shape. The remaining obstacles $\tilde{\mathcal{C}} \coloneqq \tilde{\mathcal{O}} \backslash \tilde{\mathcal{P}}$ are assumed to be strongly convex, following \cite{vasilopoulos2021reactive}, but completely unknown. As in \cite{Arslan_Koditschek_2018,vasilopoulos_pavlakos_schmeckpeper_daniilidis_koditschek_2019,vasilopoulos_pavlakos_bowman_caporale_daniilidis_pappas_koditschek_2020, vasilopoulos2021reactive}, we define the \textit{freespace} $\freespace$ as the set of collision-free placements for the closed ball $\ballclosure{\robotposition}{\robotradius}$ centered at $\robotposition$ with radius $\robotradius$. Unlike our previous work \cite{vasilopoulos2021reactive}, the positions of objects in $\tilde{\mathcal{M}}$ are also assumed to be uncertain. Instead, the robot holds a Gaussian distribution over the objects' locations, as described in Section \ref{subsec:Kalman_filter}. An illustrative example of the robot's workspace can be seen in Fig. \ref{fig:front_figure}.
The robot is equipped with a gripper to move the objects $\tilde{\mathcal{M}}$ that can either be engaged ($g=1$) or disengaged ($g=0$) and a sensor (e.g., camera) which allows it to take measurements associated with the unknown object $\tilde{\mathcal{M}}_i$. Hereafter, we assume that the robot can generate measurements as per the following \textit{observation model}:
\vspace{-1mm}
\begin{align}
    \bm{y}(t) = M(\mathbf{x}(t)) \bm{p} + \bm{v}(t)
\label{eq:observation_model}
\end{align}
\vspace{-0.3mm}
where $\bm{y}(t) = [\bm{y}_1(t)^T,\dots, \bm{y}_{N_M}^T]^T$, $\bm{y}_i(t)$ is the measurement signal at time $t$ received from position $\mathbf{x}(t)$, associated with object $\tilde{\mathcal{M}}_i$,  $\bm{p} = [\bm{p}_1^T,\dots, \bm{p}_{N_M}^T]^T, \bm{p}_i \in \reals^2$ are the objects' true positions and $\bm{v}(t) \sim \mathcal{N}(\bm{0}, \bm{R}(t))$ is a sensor-state dependent measurement noise, whose covariance matrix is $\mathbf{R}(t)$. Notice that the observation model in \eqref{eq:observation_model} is linear with respect to the objects' locations $\mathbf{p}$ but not necessarily to the robot's position $\mathbf{x}(t)$. The latter is a reasonable model for some sensors (e.g. cameras). 


\begin{assumption}\label{as:R}
The covariance matrix $\mathbf{R}(t)$ is assumed to be known for all time instants $t$, a common assumption for the application of Kalman Filters for state estimation.
\end{assumption}

\begin{figure}[t]
\captionsetup{width=\linewidth,font=footnotesize}
    \centering
    \includegraphics[width=0.48\textwidth]{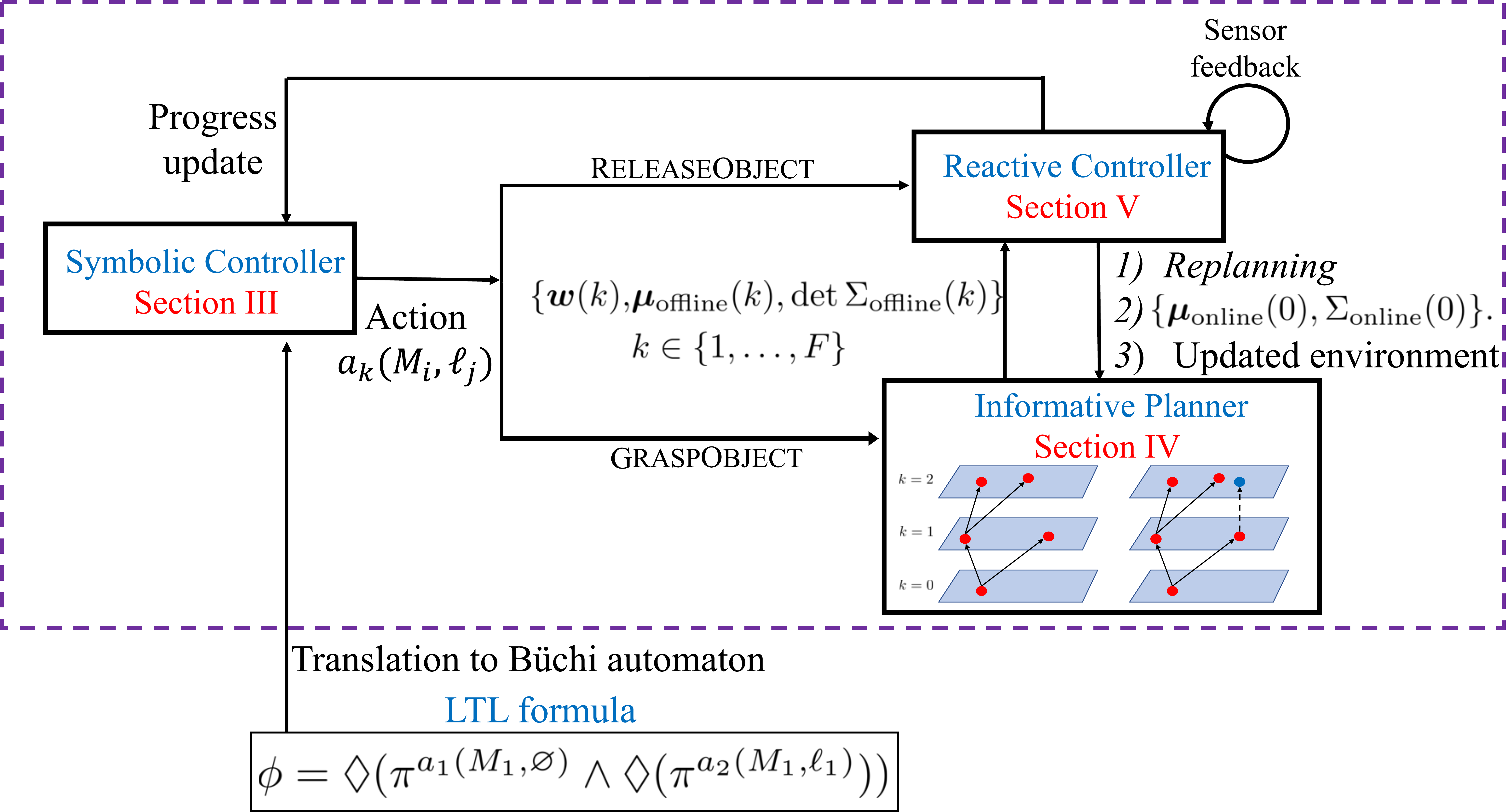}
    \setlength{\belowcaptionskip}{-18 pt}
    \caption{{\bf System architecture}: The task is encoded in an LTL formula and translated offline to a B\"uchi Automaton (Symbolic Controller - Section \ref{sec:ltl_planner}). During execution time, the Symbolic Controller generates the symbolic actions and activates either the Informative Planner (Section \ref{sec:informative_planner}) or the Reactive Controller for action implementation (Section \ref{subsec:action_implementation}). If the action is \textsc{GraspObject}, the Informative Planner gets activated to generate a sequence of waypoints $\bm{w}(k)$ that actively decrease the uncertainty of the object associated with the command. The waypoints are coupled with the statistics $(\bm{\mu}_{\text{offline}}(k), \Sigma_{\text{offline}}(k))$ of the \`{a}-posteriori Gaussian distributions, which are transmitted to the Reactive Controller for execution and get updated (using a Kalman Filter) once the robot reaches each particular waypoint. If the action is \textsc{ReleaseObject}, the Reactive Controller gets immediately activated and translates the symbolic action into a navigation command towards a target $\mathbf{x}^*$. For each action, the Reactive Controller either allows 
    the robot to probably converge to designated targets while avoiding obstacles in the environment, or switches to \textit{Replanning} mode and triggers the Informative Planner to evaluate new waypoints when specific conditions are met (Section \ref{subsec:replanning}).}
    \label{fig:system}
\end{figure}

\subsection{Kalman Filter for Position Estimation}
\label{subsec:Kalman_filter}
As mentioned above, the robot does not know ahead of time the exact locations of the movable obstacles $\mathbf{p}$. Thus, the robot must maintain a
belief over their locations and intentionally select
actions that reduce its uncertainty about the world. Given a Gaussian prior distribution for the objects' positions $\bm{p}$, i.e., $\hat{\bm{p}}(0) \sim \mathcal{N}(\bm{\mu}(0), \Sigma(0))$ and measurements denoted by $\bm{y}_{0:t}$ that the robot has collected until a time instant $t$, the robot computes the \`{a}-posteriori Gaussian distribution denoted by $\hat{\bm{p}}(t) \sim \mathcal{N}(\bm{\mu}(t\vert \bm{y}_{0:t}), \Sigma(t\vert \bm{y}_{0:t}))$, where $\bm{\mu}(t\vert \bm{y}_{0:t})$ and $\Sigma(t\vert \bm{y}_{0:t}))$ denote the \`{a}-posteriori mean and covariance matrix respectively. To compute the local Gaussian distribution, a Kalman Filter can be used.

\subsection{Specifying Complex Manipulation Tasks}
\label{subsec:specifying_tasks}
The robot needs to accomplish a mobile manipulation task, by visiting regions of interest $\ell_j\subseteq\Omega$, where $j\in\{1,\dots,L\}$, for some $L>0$, 
and applying one of the following three manipulation actions $a_k(\tilde{\mathcal{M}}_i,\ell_j)\in\ccalA$, with $\tilde{\mathcal{M}}_i \in \movableobjectsetdilated$ referring to a movable object, defined as follows:

$\bullet\textsc{GraspObject}(\tilde{\mathcal{M}}_i)$ instructing the robot to grasp the movable object $\tilde{\mathcal{M}}_i$, labeled as $a_1(\tilde{\mathcal{M}}_i, \varnothing)$, with $\varnothing$ denoting that no region is associated with this action. For the safety of the grasping operation, the robot must first localize the object. Given that the robot holds and updates a Gaussian distribution over the object's position, we can define its uncertainty over this position at time $t$ as the determinant of the \`{a}-posteriori covariance matrix, $\det\Sigma_i(t \vert \mathbf{y}_{0:t})$, where $\Sigma_i(t \vert \mathbf{y}_{0:t})$ is the covariance matrix of the marginal distribution of the joint \`{a}-posteriori Gaussian distribution, corresponding to object $\tilde{\mathcal{M}}_i$, as defined in Section \ref{subsec:Kalman_filter}. Alternative uncertainty measures could be used, such as the trace or maximum eigenvalue of $\Sigma_i(t \vert \mathbf{y}_{0:t})$. Once the robot takes the appropriate measurements at designated waypoints derived from the informative planner (Section \ref{sec:informative_planner}) and manages to reduce its uncertainty below a user-specified threshold $\epsilon$, i.e. $\det\Sigma_i(t \vert \mathbf{y}_{0:t}) \leq \epsilon$, we assume that the robot can safely grasp the object. 

$\bullet \textsc{ReleaseObject}(\tilde{\mathcal{M}}_i,\ell_j)$ instructing the robot to push the (assumed already grasped) object $\tilde{\mathcal{M}}_i$ toward its designated goal position, $\ell_j$, labeled as $a_2(\tilde{\mathcal{M}}_i,\ell_j)$.

$\bullet \textsc{DissassembleObject}(\tilde{\mathcal{M}}_j, \mathbf{x}^{**})$ gets activated by the \textit{Fix} mode, introduced in \cite[Section IV]{vasilopoulos2021reactive}, which is triggered when the completion of the $\textsc{ReleaseObject}(\tilde{\mathcal{M}}_i, \ell_j)$ is blocked by object $\tilde{\mathcal{M}}_j$. The action is responsible for moving the blocking object $\tilde{\mathcal{M}}_j$ to a different location $\mathbf{x}^{**}$.

An example of a manipulation task can be seen in Fig. \ref{fig:front_figure} where the robot is asked to grab obstacle $\tilde{\mathcal{M}}_1$ and release it in region $\ell_1$.
We will capture such manipulation tasks via Linear Temporal Logic (LTL) specifications. Specifically, we use atomic predicates of the form $\pi^{a_k(\tilde{\mathcal{M}}_i,\ell_j)}$, which are true when the robot applies the action $a_k(\tilde{\mathcal{M}}_i,\ell_j)$ and false until the robot achieves that action. Note that these atomic predicates allow us to specify temporal logic specifications defined over manipulation primitives and, unlike related works \cite{he2015towards,shoukry2018smc}, are entirely agnostic to the geometry of the environment.
We define LTL formulas by collecting such predicates in a set $\mathcal{AP}$ of atomic propositions. For example, the scenario of Fig. \ref{fig:front_figure} can be described as a sequencing task \cite{fainekos2005hybrid} with the following LTL formula: $\phi = \lozenge (\pi^{a_1(\movableobject_1,\varnothing)} \wedge \lozenge (\pi^{a_2(\movableobject_1,\ell_1)} ))$
%
where $\lozenge$ and  $\wedge$ refer to the `eventually' and `AND' operator. 
LTL formulas are satisfied over an infinite sequence of states \cite{baier2008principles}. Unlike related works where a state is defined to be the robot position, e.g., \cite{kress2009temporal}, here we follow the approach of \cite{vasilopoulos2021reactive} and define a state by the manipulation action $a_k(\tilde{\mathcal{M}}_i,\ell_j)$ that the robot applies.
Also, similarly to our prior work \cite{vasilopoulos2021reactive}, we exclude the `next' operator from the syntax, since it is not meaningful for practical robotics applications \cite{kloetzer2008fully}, as well as the negation operator\footnote{Since the negation operator is excluded, safety requirements, such as obstacle avoidance, cannot be captured by the LTL formula; nevertheless, the proposed method can still handle safety constraints by construction of the (continuous-time) reactive, vector field controller in Section~\ref{sec:reactive_planner}.}.

\begin{remark}
The robot does not know where the objects are exactly, before grasping them. However, if it manages to grasp an object and releases it, it will remember the last location of the released obstacle. 
\end{remark}

\subsection{Problem Statement}
Given a robot of initial rigid placement $\robotpositionunicycle(0)$, equipped with a sensor (e.g., camera) capable of receiving noisy measurements as per the observation model \eqref{eq:observation_model} and a prior Gaussian distribution $\hat{\mathbf{p}}(0)$ over the objects' positions $\mathbf{p}$, design a hybrid control architecture that (i) computes in real time a set of informative waypoints  $\mathcal{W}$, that will allow the robot to actively reduce its uncertainty over the positions $\mathbf{p}$, and (ii) designs an infinite sequence of symbolic/discrete manipulation actions that satisfy $\phi$ and a continuous-time controller to execute it while avoiding all obstacles.



\subsection{Hybrid Controller Architecture}
In this Section, we briefly provide an overview of our proposed hybrid controller architecture, seen also in Fig.~\ref{fig:system}.
The manipulation task, expressed in $\phi$, is fed to the symbolic controller (Section \ref{sec:ltl_planner}) which translates it online into a sequence of symbolic actions. The robot nominally executes each of these actions sequentially. The informative controller (Section \ref{sec:informative_planner}) is called to compute a sequence of waypoints that if followed, active reduction of the uncertainty over the objects' positions is guaranteed. The reactive controller (Section \ref{sec:reactive_planner}) receives the waypoints, follows them by avoiding obstacles in the unknown environment and activates, under special occasions, the \textit{Replanning} mode.  

%% file: 3-ltl-planner-icra.tex
\section{SYMBOLIC CONTROLLER}
\label{sec:ltl_planner}
%

In this Section we present the discrete controller that generates online a sequence of manipulation commands in the form of actions as described in Section \ref{sec:problem_formulation}. A detailed construction of the controller can be found in \cite[Sec. III]{vasilopoulos2021reactive}.



\subsection{Construction of the Symbolic Controller}\label{sec:nba}

Initially, the LTL specification $\phi$, constructed using a set of atomic predicates $\mathcal{AP}$, is translated into a Non-deterministic B$\ddot{\text{u}}$chi Automaton (NBA)  using the tool in \cite{gastin2001fast}. The NBA contains state-space and transitions among states. The LTL formula is satisfied if starting from the initial state, the robot generates an infinite sequence of observations (i.e., atomic predicates that become true) that results in an infinite sequence of transitions so that the final state is visited infinitely often. The NBA states can be used to measure the progress the robot has made in terms of accomplishing the assigned mission, using a distance metric described in \cite[App. I]{vasilopoulos2021reactive}. Specifically given that the robot lies at a specific NBA state $\mathbf{s}_B(t)$, the metric yields the next state $\mathbf{s}_B^{\text{next}}$ that decreases the distance to a state that accomplishes the assigned task. Once the target NBA state is selected, a symbolic action in the form of a manipulation task that achieves it is generated and navigation commands are passed either to the informative planner or to the reactive controller.

When the symbolic action is satisfied, a new target NBA state is selected and a corresponding manipulation command is generated. In case of incompleteness, (e.g. the reactive controller can not reach an object) the controller picks another symbolic action that could drive the robot from the state $\mathbf{s}_B(t)$ to the selected state $\mathbf{s}_B^{\text{next}}$. If no commands exist, the symbolic controller picks a different state $\mathbf{s}_B^{\text{next}}$ that could drive the robot towards the final NBA state. If there are no such automaton states, the symbolic controller returns a message stating that the robot cannot accomplish the task.

%% file: 5-informative-sampling-based-planner.tex
\section{INFORMATIVE PLANNER}
\label{sec:informative_planner}
In this Section, we present a sampling-based path planner \cite{tzes2021distributed}
that generates informative waypoints so that the uncertainty of the object of interest drops below a user-specified threshold; see Section \ref{subsec:specifying_tasks}. The informative planner gets activated each time a new grasping command is received from the symbolic controller or the reactive controller turns on the \textit{Replanning} mode; see Section \ref{subsec:replanning}. The robot temporarily pauses and waits for the transmission of new waypoints from the informative planner, before resuming the execution of the action using the reactive controller.

\subsection{Construction of the Informative Path Planner}
The path planner receives from the symbolic controller the action $\textsc{GraspObject}(\tilde{\mathcal{M}}_i) $ and translates it into a navigation command towards a collision-free location $\mathbf{x}^*$ on the boundary of the closed ball $\ballclosure{\bm{\mu}_{i,\text{offline}}(0)}{\robotradius+\movableobjectradius{i}}$, where $\bm{\mu}_{i, \text{offline}}(0)$ is the mean of the marginal distribution of the joint prior Gaussian distribution $\hat{\mathbf{p}}(0) \sim \mathcal{N}(\bm{\mu}_{\text{offline}}(0), \Sigma_{\text{offline}}(0))$ (see Section \ref{subsec:Kalman_filter}) when the informative planner is called. The task of the planner is to design an informative path that would drive the robot towards the target $\mathbf{x}^*$ and allow it to take measurements to reduce the uncertainty it has over the object's position. Given the initial robot's position $\mathbf{x}(0)$ and a prior distribution of the objects positions $\mathbf{p}(0)$, our goal is to compute a planning horizon $F$ and a sequence of waypoints $\bm{w}(k) \in \mathbb{R}^2$, for all time instants $k=\{0,\dots,F\}$, which solves the deterministic optimal control problem in \eqref{eq:Prob1},
where $\bm{\delta}(k) \in \mathcal{U}$ is an actuation input selected from a finite set $\mathcal{U}$ of admissible actuation inputs. In \eqref{eq:obj1}, $\bm{w}_{0:F}$ stands for the sequence of waypoints from $k=0$ until $k=F$. The objective \eqref{eq:obj1} captures the cumulative uncertainty in the estimation of $\mathbf{p}_i(k)$ after fusing information collected by the robot from $k=0$ up to time $F$. The constraint \eqref{eq:constr11} requires the final uncertainty of $\mathbf{p}_i(F)$ to be below a user-specified threshold $\epsilon$ and the second constraint \eqref{eq:constr12} requires that the returned waypoints should lie in obstacle-free areas. The constraint \eqref{eq:constr13} computes the following waypoint after applying the actuation input $\bm{\delta}(k)$ and constraint \eqref{eq:constr14} sets the first waypoint equal to the initial position of the robot. In \eqref{eq:constr15}, $\xi(\cdot)$ stands for the Kalman Filter Ricatti map used to compute the covariance matrices given the robot state\footnote{The initial robot position and prior distribution refer to the time instant where the informative planner was called. We use $k$ to denote the offline time instances and distinguish them from the online time instances $t$.}.
\vspace{-1mm}
\begin{subequations}
\label{eq:Prob1}
\begin{align}
& \min_{\substack{ F, \bm{w}_{0:F}}} \left[J(F,\bm{w}_{0:F}) = \sum_{k=0}^{F}  \det\Sigma_{i,\text{offline}}(k+1) \right] \label{eq:obj1}\\
& \ \ \ \ \ \ \ \det\Sigma_{i,\text{offline}}(F+1)\leq \epsilon \label{eq:constr11} \\
& \ \ \ \ \ \ \ \    \bm{w}(k+1) \in \mathcal{F}, \label{eq:constr12} \\
& \ \ \ \ \ \ \ \  \bm{w}(k+1) = \bm{w}(k) + \bm{\delta}(k) , \label{eq:constr13}\\
& \ \ \ \ \ \ \ \ \bm{w}(0) = \mathbf{x}(0), \label{eq:constr14}\\
& \ \ \ \ \ \ \ \ \Sigma_{\text{offline}}(k+1) =\xi(\bm{w}(k),\Sigma_{\text{offline}}(k))\label{eq:constr15}
\end{align}
\end{subequations} 


To solve \eqref{eq:Prob1} we employ a sampling-based algorithm that incrementally constructs a directed tree that explores both the information and the physical space. To what follows, we denote the tree as $\mathcal{G}=\{\mathcal{V}, \mathcal{E}, J_{\mathcal{G}} \}$, where $\mathcal{V}$ is the set of nodes and $\mathcal{E} \subseteq \mathcal{V} \times \mathcal{V}$ denotes the set of edges. Each node of the tree contains states of the form $\mathbf{q}(k) = [\mathbf{x}(k), \Sigma_{\text{offline}}(k)]$ and the function $J_{\mathcal{G}} \colon \mathcal{V} \rightarrow \mathbb{R}_{+}$ assigns the cost of reaching node $\mathbf{q}(k) \in \mathcal{V}$ from the root of the tree. The root of the tree is defined as $\mathbf{q}(0) = [\mathbf{x}(0), \Sigma_{\text{offline}}(0)]$, where $\mathbf{x}(0), \Sigma_{\text{offline}}(0)$ are the robot's initial position and prior covariance respectively. The cost of the root is initialized as $J_{\mathcal{G}}(\mathbf{q}(0)) = \det \Sigma_{\text{offline}}(0)$, while the cost of node $\mathbf{q}(k+1)$ is equal to $J_{\mathcal{G}}(\mathbf{q}(k+1)) = J_{\mathcal{G}}(\mathbf{q}(k)) + \det \Sigma_{\text{offline}}(k+1)$, where the node $\mathbf{q}(k)$ is the parent node of $\mathbf{q}(k+1)$. Applying the cost function recursively results in the objective function \eqref{eq:obj1}. A detailed description for the construction of the tree $\mathcal{G}$ is provided in \cite[Section III]{tzes2021distributed}. Except for the waypoints $\bm{w} \in \mathcal{W}$, the informative planner sends their corresponding mean $\bm{\mu}_{\text{offline}}$ and covariance matrix $\Sigma_{\text{offline}}$ from the \`{a}-posteriori gaussian distribution, computed during the design of the informative path, i.e. the set $\{\bm{w}(k), \bm{\mu}_{\text{offline}}(k), \det\Sigma_{\text{offline}}(k) \},\ k \in \{1,\dots, F\}$. Once a sequence of informative waypoints has been generated, we append to the end of this sequence the target $\mathbf{x}^*$ for grasping. 


\begin{theorem}[\textbf{Probabilistic Completeness} \cite{tzes2021distributed}]\label{thm:probCompl}
\textit{If there exists a solution to Problem \ref{eq:Prob1}, then the Informative Planner is probabilistically complete, i.e., feasible waypoints $\bm{w}_{0:F} = \bm{w}(0), \dots, \bm{w}(F)$ will be found with probability $1$.}
\end{theorem}


\begin{remark}
The optimal control problem \eqref{eq:Prob1} has resulted from a stochastic optimal control problem discussed in \cite[Section II]{tzes2021distributed} that depends on sensor measurements. Then, due to the linearity and Gaussian assumptions made about the sensor model \eqref{eq:observation_model}, a separation principle presented in \cite{6907811} is employed, which allows the conversion of the stochastic optimal control problem into the deterministic one shown in \eqref{eq:Prob1} that does not depend on the measurements $\mathbf{y}_{0:k}$. 
Therefore, each time informative paths are required to decrease the uncertainty of an object, the robot can pause and compute the informative waypoints as per \eqref{eq:Prob1} without the need of measurements. To emphasize this, in \eqref{eq:Prob1}, we use the subscript `offline' in the computed covariance matrices.
\end{remark}
\vspace{-3mm}
\begin{remark}
The formulation of \eqref{eq:Prob1} requires the covariance matrix $\Sigma_{\text{offline}}$ 
 to be updated using the Kalman Filter update rule. Since the update rule and in general the entire problem does not depend on the measurements $\mathbf{y}_{0:k}$, the covariance matrix $\Sigma_{\text{offline}}$ gets updated offline. However, the mean $\bm{\mu}_{\text{offline}}(k)$ is not updated during the computation of the informative waypoints and remains equal to the prior mean $\bm{\mu}_{\text{offline}}(0)$ for all time instances $k$. On the contrary, during online execution, each time the robot reaches a waypoint $\bm{w}(k)$, the reactive controller will update both the mean $\bm{\mu}_{\text{online}}(k+1 \vert \mathbf{y}_{0:k})$ and the covariance matrix $\Sigma_{\text{online}}(k+1 \vert \mathbf{y}_{0:k})$ based on the received measurements; see Section \ref{sec:reactive_planner}.
\end{remark}

%% file: 6-reactive-planner.tex
\section{REACTIVE CONTROLLER}
\label{sec:reactive_planner}

In this Section, we briefly describe the continuous, reactive controller \cite{vasilopoulos_pavlakos_bowman_caporale_daniilidis_pappas_koditschek_2020}, responsible for driving the robot to track either a single target $\mathbf{x}^*$ or a set of waypoints $\mathcal{W}$. The reactive controller receives from the symbolic controller the action $\textsc{ReleaseObject}(\tilde{\mathcal{M}}_i, \ell_j)$ and translates the action into a navigation command towards target $\mathbf{x}^*$, selected as the centroid of region $\ell_j$. Upon receiving a symbolic action $\textsc{GraspObject}(\tilde{\mathcal{M}}_i)$, the reactive controller first awaits from the informative planner a set of waypoints $\mathcal{W}$, and then sequentially drives the robot to each target-waypoint $\bm{w} \in \mathcal{W}$. When the robot reaches the final waypoint, the reactive controller creates the gripping command $g=1$, and requests a new action from the symbolic controller. Next, we briefly describe the online implementation of the controller, and then analyze the conditions under which the reactive controller triggers \textit{replanning} from the informative planner.

\subsection{Action Implementation}
\label{subsec:action_implementation}
The reactive vector field controller from our prior work \cite{vasilopoulos_pavlakos_bowman_caporale_daniilidis_pappas_koditschek_2020} allows either a fully-actuated or a differential-drive robot to probably converge to a designated fixed target while avoiding obstacles in the environment. Specifically, the known components of the environment (e.g., walls) and any sensed fragments of the unknown obstacles are stored in the mapped space from which a change of coordinates $\mathbf{h}$ deforms the mapped space to yield a geometrically simple but topologically equivalent model space. A constructed vector field in this model space is then transformed in realtime through the diffeomorphism $\mathbf{h}$ to generate the input in the physical space. When the robot grips an object, we use the method from \cite{vasilopoulos2018}, where the vector-field controller designs control policies $\mathbf{u}_{i,c}$ for the center $\mathbf{x}_{i,c}$ of the circumscribed disk of radius $(\rho_i + r)$, enclosing the robot and the object. The inputs are then transformed into the differential drive commands $\bar{\mathbf{u}}=(v,\omega)$ using the Jacobian $\mathbf{T}_{i,c}(\psi)$ of the gripping contact, i.e., $\bar{\mathbf{u}}\coloneqq \mathbf{T}_{i,c}(\psi)^{-1}\mathbf{u}_{i,c}, \  \dot{\mathbf{x}}_{i,c} = \mathbf{T}_{i,c}(\psi) \bar{\mathbf{u}}$. 

\begin{figure}[t]
\captionsetup{width=\linewidth,font=footnotesize}
     \centering  
     \begin{subfigure}[b]{0.2\textwidth}
         \centering
         \includegraphics[width=\textwidth]{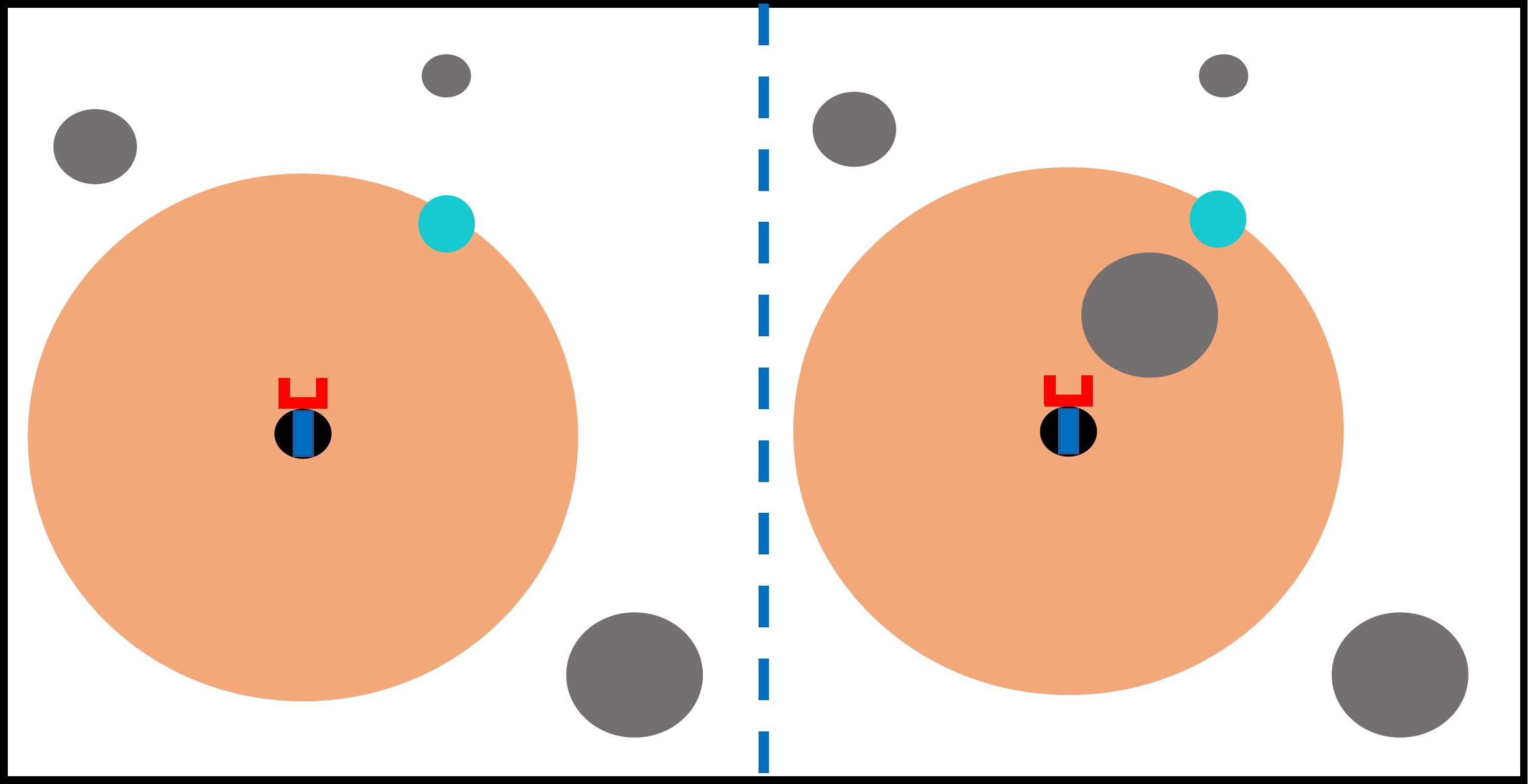}
         \caption{}
       \label{fig:replanning_1}
     \end{subfigure}
          \hfill
\begin{subfigure}[b]{0.2\textwidth}
         \centering
     \includegraphics[width=\textwidth]{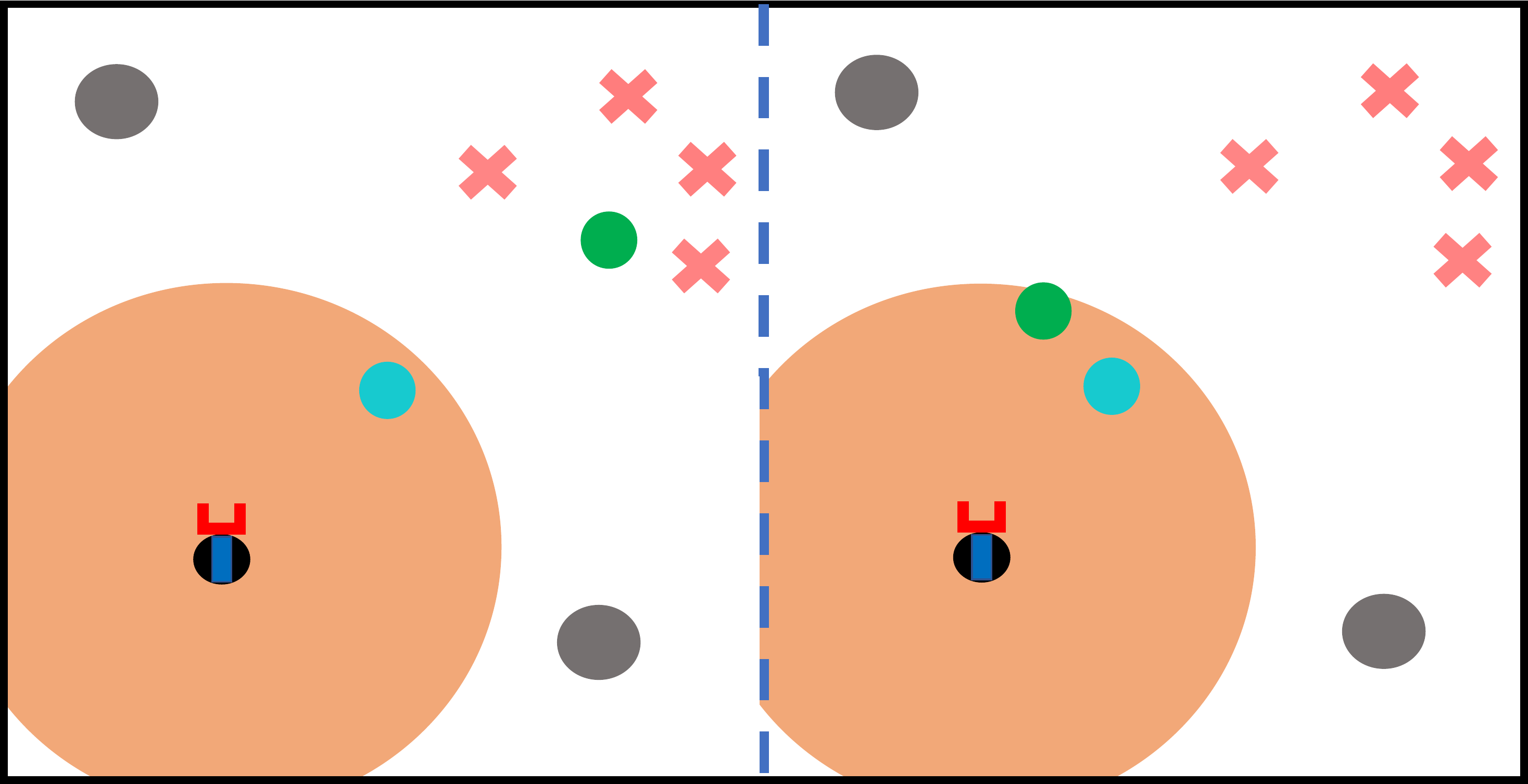}
         \caption{}
       \label{fig:replanning_2}
     \end{subfigure}
          \vfill
     \begin{subfigure}[b]{0.2\textwidth}
         \centering
       \includegraphics[width=\textwidth]{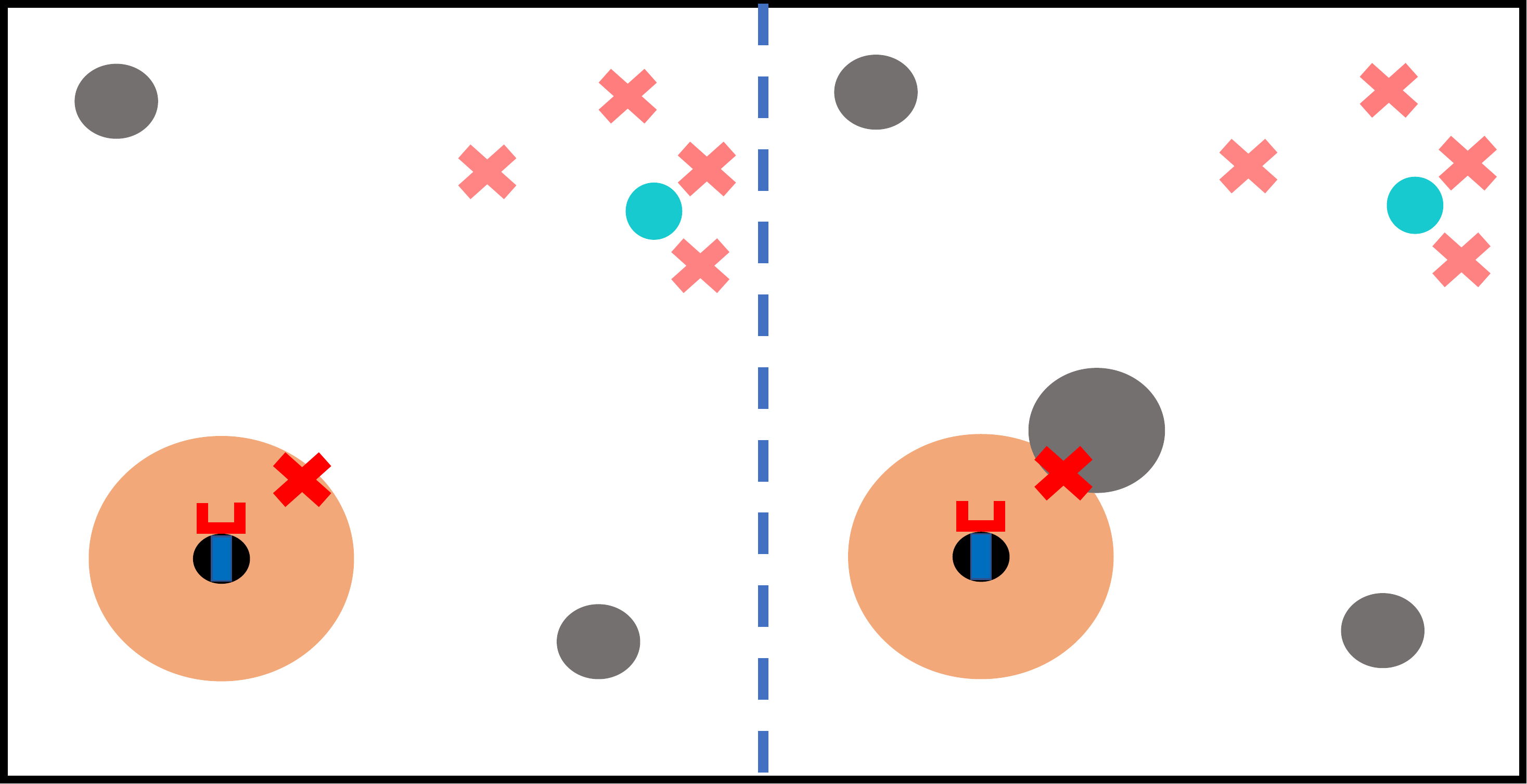}
         \caption{} \label{fig:replanning_3}
     \end{subfigure}
          
     \setlength{\belowcaptionskip}{-18 pt}
     \caption{Figs. \ref{fig:replanning_1}-\ref{fig:replanning_3} represent examples of the three replanning conditions described in Section \ref{subsec:replanning}. On the left and right subparts of the figures we include examples of the estimated and actual configutations of the world, respectively. In Fig. \ref{fig:replanning_1} the informative planner was not aware of the existence of the gray obstacle and thus expected to sense the object and returned a smaller uncertainty than the one computed from the reactive controller. In Fig. \ref{fig:replanning_2}, the estimated position of the object is depicted in green and there is a large deviation between the informative planner's and the reactive controller's expected locations, $\bm{\mu}_{i,\text{offline}}, \bm{\mu}_{i,\text{online}} $ respectively. Finally, Fig. \ref{fig:replanning_3} visualizes the case of an invalid waypoint.}

     \label{fig:replanning_examples}
\end{figure}

\subsection{Replanning conditions}
\label{subsec:replanning}
The reactive controller is further responsible for detecting whether or not a replanning request should be forwarded to the informative planner. As mentioned in Section \ref{sec:informative_planner}, the reactive controller receives the set $\{\bm{w}(k), \bm{\mu}(k), \det\Sigma(k) \},\ k \in \{1,\dots, F\}$. Before the online execution of the path, the robot sets its prior gaussian distribution equal to the prior distribution of the informative planner, i.e., $\bm{\mu}_{\text{online}}(0)=\bm{\mu}_{\text{offline}}(0), \Sigma_{\text{online}}(0) = \Sigma_{\text{offline}}(0)$. When the robot reaches waypoint $\bm{w}(k)$, driven by the reactive controller, it updates the gaussian distribution using Kalman Filter and measurements received at each one of the previous $k-1$ waypoints, and computes the \`{a}-posteriori mean $\bm{\mu}_{\text{online}}(k\vert \mathbf{y}_{0:k})$ and covariance matrix $\Sigma_{\text{online}}(k \vert \mathbf{y}_{0:k})$. Using this information, the \textit{Replanning} mode gets activated if at waypoint $\bm{w}(k)$: (i) the robot expected smaller uncertainty for object $\tilde{\mathcal{M}}_i$, i.e., $\det\Sigma_{i, \text{online}}(k \vert \mathbf{y}_{0:k}) > \det\Sigma_{i, \text{offline}}(k) + \epsilon_{\Sigma}$, for some small $\epsilon_{\Sigma}>0$, (ii) the estimated position of the object $\tilde{\mathcal{M}}_i$ deviates from the offline estimation, i.e., $\parallel \bm{\mu}_{i,\text{online}}(k\vert \mathbf{y}_{0:k}) - \bm{\mu}_{i,\text{offline}}(k) \parallel_2 > \epsilon_{\mu}$, for some small $\epsilon_{\mu}>0$, and (iii) the upcoming waypoint, $\bm{w}(k+1)$, lies inside an initially unknown but now sensed obstacle. Illustrative examples of the replanning conditions can be seen in Fig. \ref{fig:replanning_1}-\ref{fig:replanning_3}. 
If the \textit{Replanning} mode is activated, the reactive controller transfers its latest estimation about the location of the objects to the informative planner, which uses it as its prior distribution and the updated environment.

%% file: 7-simulations.tex
\section{ILLUSTRATIVE SIMULATIONS}
\label{sec:simulations}

In this Section, we present numerical experiments that illustrate the performance of our proposed architecture for rearrangement planning (see Fig. \ref{fig:system}). All case studies have been implemented using Python 3.7 (Informative Planner and Reactive Controller) and MATLAB 2020a (Symbolic Controller), on a Macbook Pro, 2.7 GHz Quad-Core Intel Core i7, 16GB RAM. 
In each case, we assume that the robot can take noisy measurements with a camera that provides the $xy$-coordinates of any {\it visible} movable object\footnote{For simulation purposes, we consider an object to be visible if the straight-line path from the robot to the true object location is collision-free.} within its sensing range, expressed in the robot's frame. The additive, measurement noise is Gaussian as defined in the observation model \eqref{eq:observation_model}, where $\mathbf{R}=(0.05\odot\mathbf{p})^2\mathbb{I}$. The robot perfectly localizes object $\tilde{\mathcal{M}}_i$ when the uncertainty drops below threshold $\epsilon$, set to $1e^{-3}$. Finally, the \textit{Replanning} thresholds $\epsilon_{\Sigma} ,\epsilon_{\bm{\mu}}$ are set as $1e^{-3}$ and $0.3m$ respectively.

\subsection{Rearrangement Task}
\begin{figure}[t]
\captionsetup{width=\linewidth,font=footnotesize}
    \centering
    \includegraphics[width=0.45\textwidth]{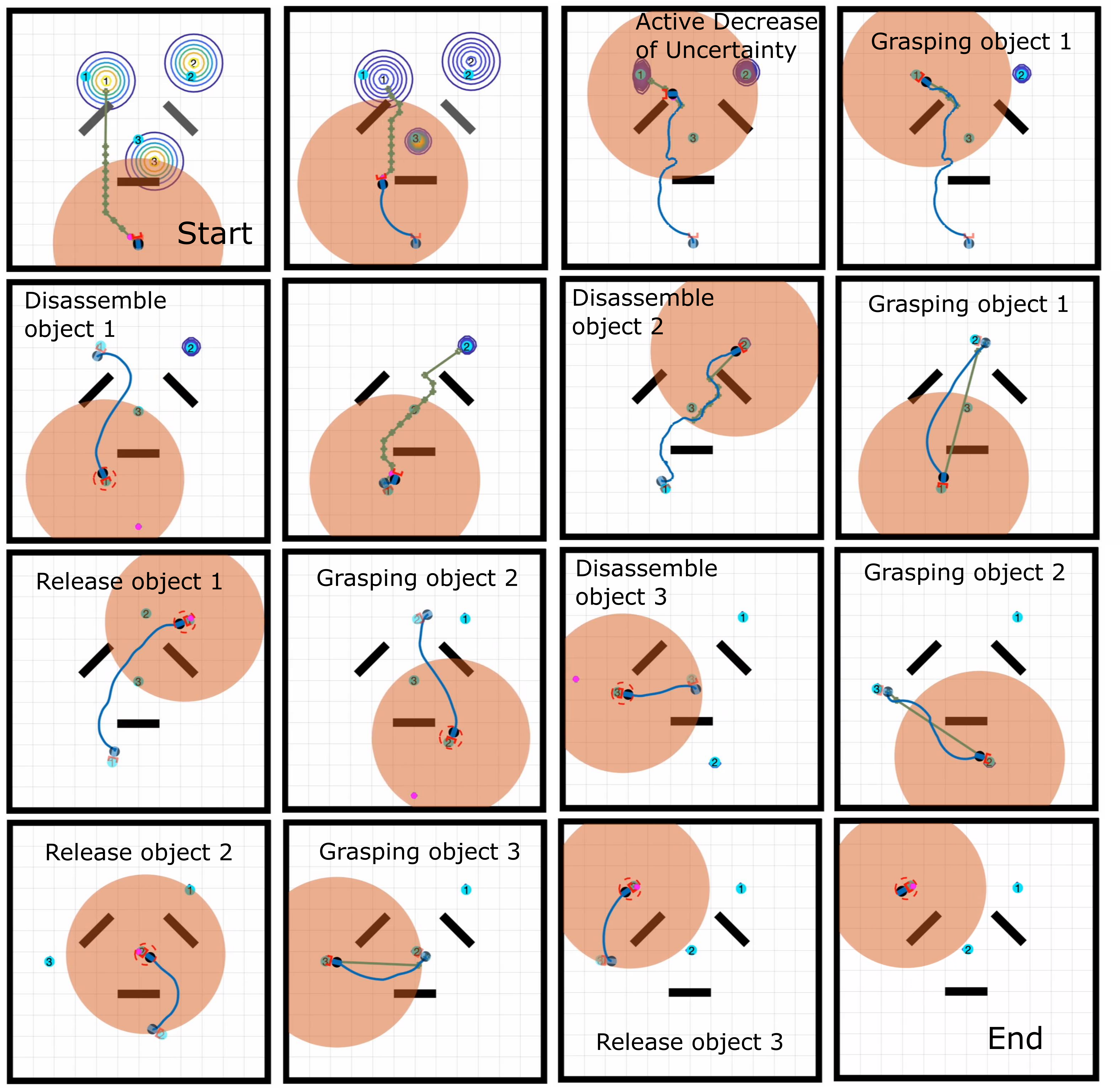}
    \setlength{\belowcaptionskip}{-18 pt}
    \caption{An illustrative execution of the clockwise rearrangement of three movable objects (cyan), in an environment cluttered with some unanticipated obstacles (initially dark grey and then black upon sensed and localized).}
    \label{fig:rearrangement_4x4}
\end{figure}

We applied our proposed architecture in a complex rearrangement task, where the robot is required to rotate clockwise three movable objects. In this case, the reactive controller activates the \textit{Fix} mode and executes the $\textsc{DissassembleObject}(\tilde{\mathcal{M}}_j, \mathbf{x}^{**})$ action, invisible to the symbolic controller, before resuming the execution of the initially received action. Fig. \ref{fig:rearrangement_4x4} illustrates successive snapshots of the rearrangement planning scenario. At each figure, we show the marginal \`{a}-posteriori gaussian  distributions for the objects' positions, as computed by the reactive controller, using the online Kalman Filter. Initially, the robot follows the informative waypoints (green) to execute $\textsc{GraspObject}(1)$. During its movement, it actively reduces the uncertainty it has over the objects' locations $\mathbf{p}$, while the mean $\bm{\mu}_{1, \text{online}}$ converges to the actual position $\mathbf{p}_1$. Next, the \textit{Fix} mode gets activated, where the robot executes sequentially the actions $\textsc{DissassembleObject}(1,\mathbf{x}^{**})$, and $\textsc{DissassembleObject}(2,\mathbf{x}^{***})$. Finally, the robot resumes the action $\textsc{ReleaseObject}(1,2)$ and follows a similar procedure for the placement of objects 2 and 3.

\begin{figure}[t]
\captionsetup{width=\linewidth,font=footnotesize}
     \centering      
     \begin{subfigure}[b]{0.23\textwidth}
         \centering
         \includegraphics[width=\textwidth]{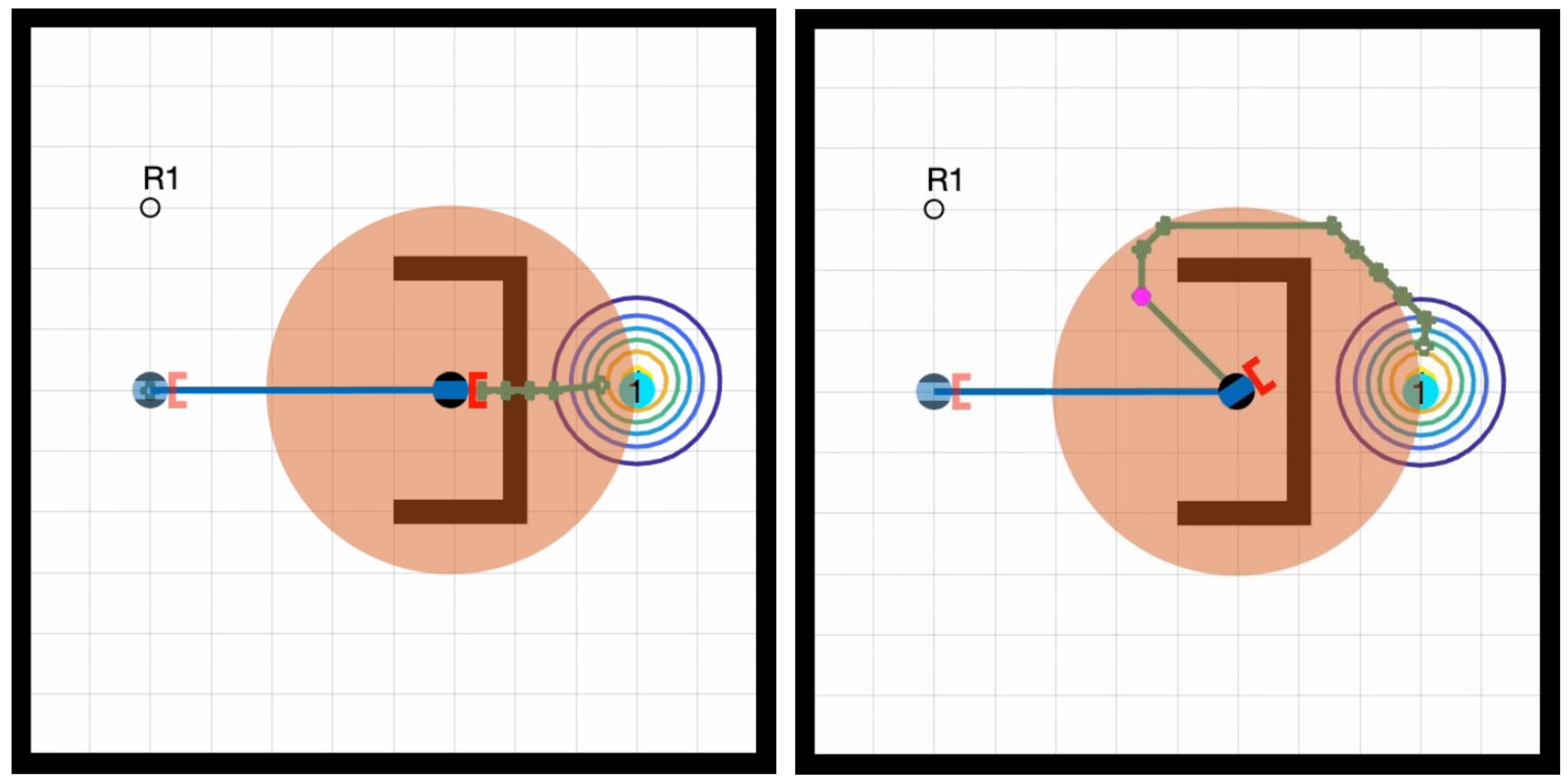}
         \caption{}
     \label{fig:sim_replanning_1}
     \end{subfigure}
     \hfill      
     \begin{subfigure}[b]{0.23\textwidth}
         \centering
         \includegraphics[width=\textwidth]{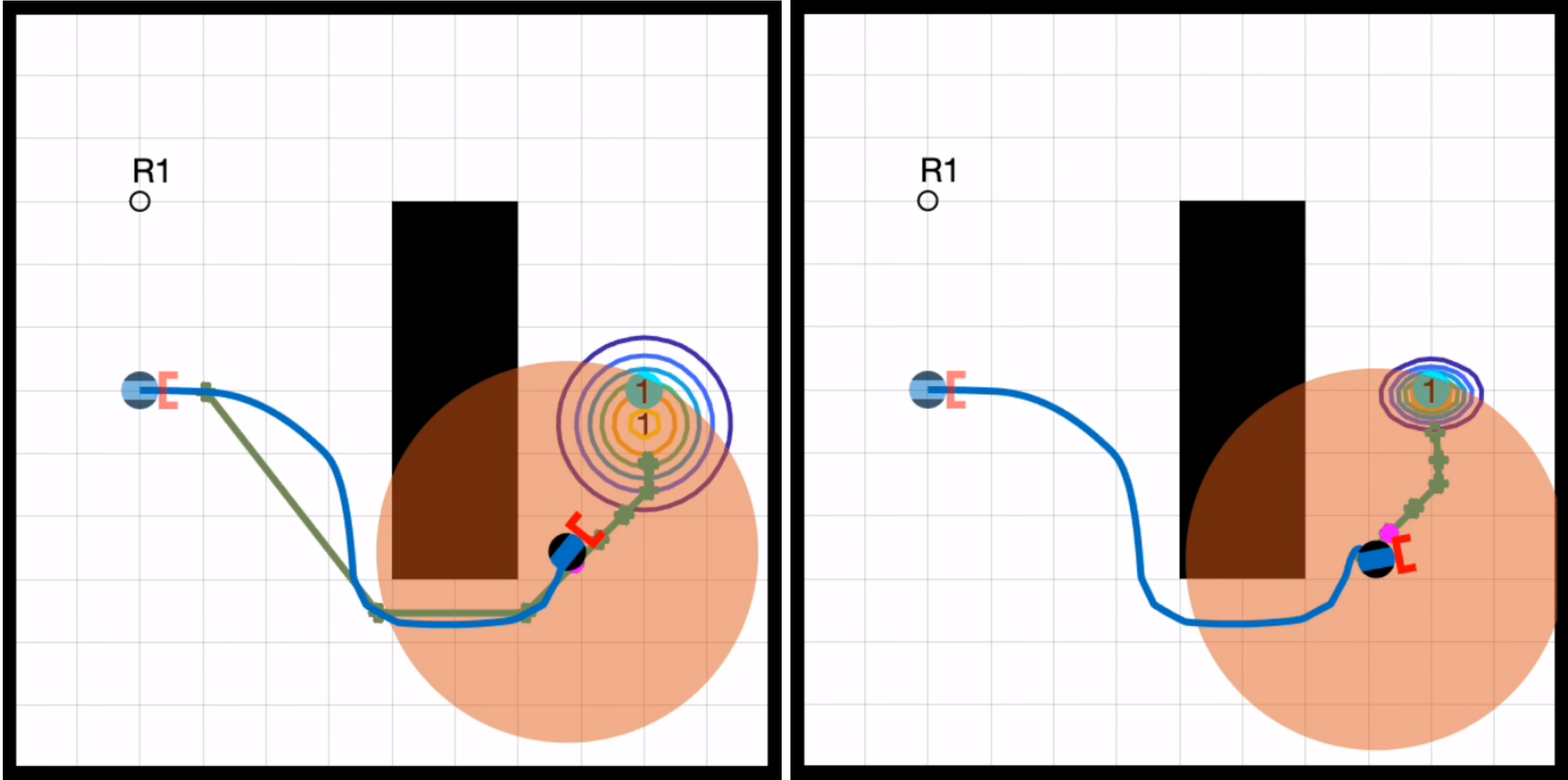}
         \caption{}
       \label{fig:sim_replanning_2}
     \end{subfigure}
          \vfill
    \begin{subfigure}[b]{0.23\textwidth}
         \centering
     \includegraphics[width=\textwidth]{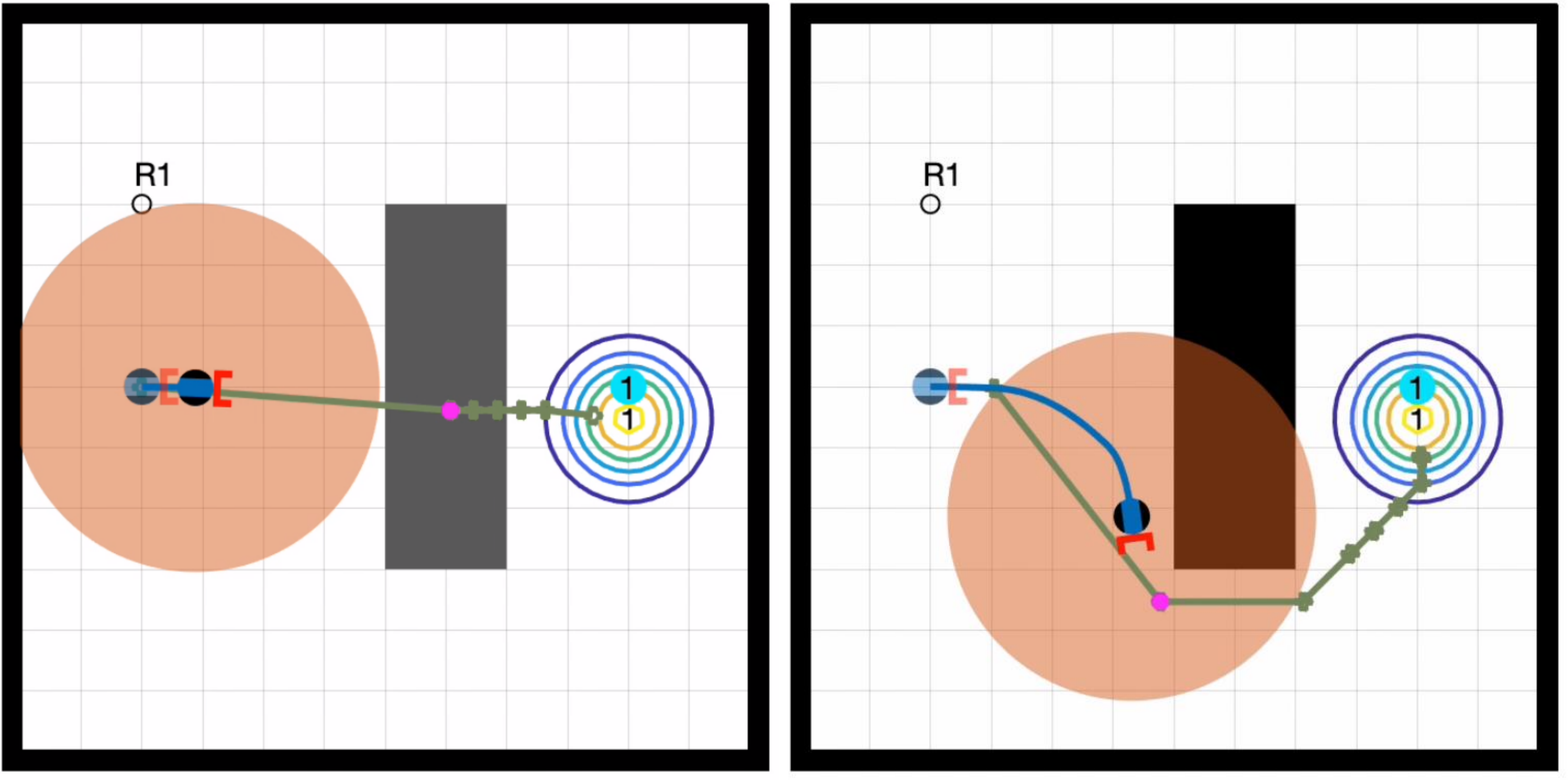}
         \caption{}
       \label{fig:sim_replanning_3}
     \end{subfigure}
     \setlength{\belowcaptionskip}{-13 pt}
     \caption{Illustrative examples that lead to triggering the \textit{Replanning} mode, described in Section \ref{subsec:replanning}.}
     \label{fig:replanning_simulations}
\end{figure}

\subsection{Demonstration of Replanning Scenarios}
In this Section, we present simulation examples that trigger the \textit{Replanning} mode, in the simple scenario where the robot is tasked with repositioning a movable object to a predefined location (labeled R1 in Fig. \ref{fig:replanning_simulations}). On the left subparts of Fig. \ref{fig:replanning_simulations}, we present the state of the world just before the \textit{Replanning} gets triggered and on the right the updated informative waypoints. In Fig. \ref{fig:sim_replanning_1}, an initially unknown obstacle lays between the robot and the expected location of the object. Initially, the informative planner wrongly assumes that the object can be sensed, and thus the uncertainty $\det \Sigma_{\text{offline}}$ is expected to decrease. However, during online execution, the object's visibility is blocked by the obstacle, and thus its uncertainty $\det \Sigma_{\text{online}}$ is larger than expected, i.e., Assumption \ref{as:R} is violated due to unknown obstacles. In Fig. \ref{fig:sim_replanning_2}, a replan is triggered because the expected location of the object $\bm{\mu}_{\text{online}}$ is much different than the one computed offline, $\bm{\mu}_{\text{offline}}$, making grasping of the object unsafe. Finally, in Fig. \ref{fig:sim_replanning_3} the waypoint to be tracked (magenta) is infeasible. The reactive controller senses the unanticipated obstacle, the informative planner's world gets updated and valid waypoints are recomputed.
%

%
%
%

\begin{figure}[t]
\captionsetup{width=\linewidth,font=footnotesize}
    \centering
    \includegraphics[width=0.35\textwidth]{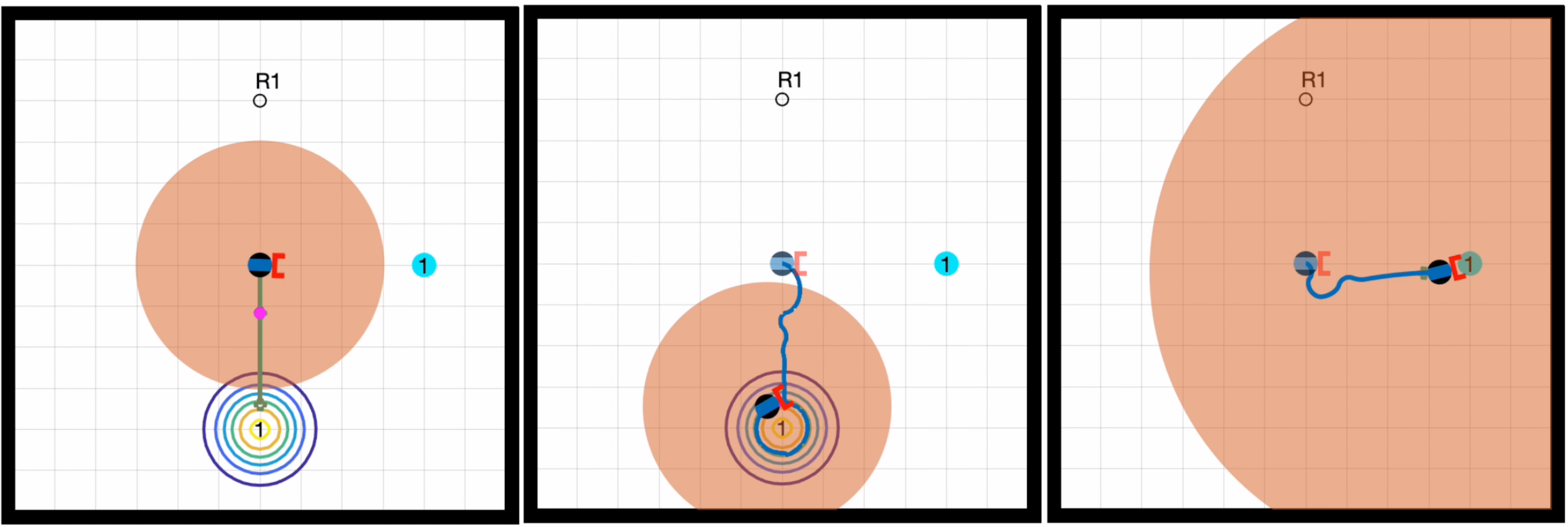}
     \setlength{\belowcaptionskip}{-18 pt}
    \caption{Illustration of an extreme case where the prior estimated location of the object is far away from the true object location (Left). The architecture reports failure for a small sensing range (Middle) but succeeds with a larger range (Right).}
    \label{fig:sim_extreme}
\end{figure}

\subsection{Demonstration of Failure Mode}
Finally, in Fig \ref{fig:sim_extreme}, we report an extreme case where our architecture fails. The prior expected location, $\bm{\mu}_\text{offline}$ (center of the Gaussian) is far away from the actual object position (cyan), as can be seen in the left subfigure. In the middle subfigure, the robot has a small sensing range and cannot take a measurement from the object in order to call for $\textit{Replanning}$, thus ending up detouring from the target. The larger sensing range in the right subfigure allows the robot to replan and complete its action.


%% file: 8-conclusion.tex
\section{CONCLUSION}
\label{sec:conclusion}
In this paper, we introduce a novel hybrid architecture for rearrangement planning under sensing and environmental uncertainty, that can design informative paths for \textit{actively} reducing the uncertainty a \textit{mobile} robotic manipulator maintains over the task domain, while executing complex manipulation tasks. Future work will focus on extending the presented architecture to multi-robot systems performing collaborative manipulation tasks as well as on experimental validation.